\documentclass[10pt]{article} %
\usepackage[preprint]{tmlr}

\usepackage{amsmath,amsfonts,bm}

\def\eqref#1{equation~\ref{#1}}

\def\1{\bm{1}}

\DeclareMathAlphabet{\mathsfit}{\encodingdefault}{\sfdefault}{m}{sl}
\SetMathAlphabet{\mathsfit}{bold}{\encodingdefault}{\sfdefault}{bx}{n}

\usepackage{hyperref}
\usepackage{url}
\usepackage{graphicx}
\usepackage{listings}
\usepackage{color}
\usepackage{multirow}
\usepackage{array}
\usepackage{textgreek}
\usepackage{longtable}
\usepackage{caption}
\usepackage{booktabs}
\usepackage{subcaption}

\newcolumntype{C}[1]{>{\centering\let\newline\\\arraybackslash\hspace{0pt}}m{#1}}
\newcolumntype{L}[1]{>{\raggedright\let\newline\\\arraybackslash\hspace{0pt}}m{#1}}
\newcolumntype{R}[1]{>{\raggedleft\let\newline\\\arraybackslash\hspace{0pt}}m{#1}}

\definecolor{deepblue}{rgb}{0,0,0.7}
\definecolor{deepred}{rgb}{0.7,0,0}
\definecolor{deepgreen}{rgb}{0,0.5,0}

\newcommand\pythonstyle{\lstset{
    language=Python,
    basicstyle=\ttfamily,
    morekeywords={},              %
    keywordstyle=\ttfamily\color{deepblue},
    emph={get_n_params,get_flops_per_seq}, %
    emphstyle=\ttfamily\color{deepred},    %
    commentstyle=\color{deepgreen},
    stringstyle=\color{deepgreen},
    frame=tb,                         %
    showstringspaces=false
}}

\lstnewenvironment{python}[1][] {
    \pythonstyle
    \lstset{#1}
} {}

\makeatletter
\newcommand{\thickhline}{%
    \noalign {\ifnum 0=`}\fi \hrule height 1pt
    \futurelet \reserved@a \@xhline
}
\makeatother

\title{\centering Position Interpolation Improves ALiBi Extrapolation}
\author{
    \centering
    \name Faisal Al-Khateeb, Nolan Dey, Daria Soboleva, Joel Hestness \\ \vspace{6pt} %
    \affil Cerebras Systems~~~~
    \email faisal.alkhateeb@cerebras.net
}

\begin{document}

\maketitle
\begin{abstract}
Linear position interpolation helps pre-trained models using rotary position embeddings (RoPE) to extrapolate to longer sequence lengths. We propose using linear position interpolation to extend the extrapolation range of models using Attention with Linear Biases (ALiBi). We find position interpolation significantly improves extrapolation capability on upstream language modelling and downstream summarization and retrieval tasks.
\end{abstract}

\section{Introduction}
LLMs support for long sequence lengths is critical for many downstream applications. However, retraining or fine-tuning models to support long sequence lengths is costly. Recent works by \cite{dosovitskiy-imageTransf} and \cite{chen-pi} show approaches to interpolate learned and rotary position embeddings (RoPE), respectively. These techniques improve models' capability to interpolate and extrapolate to different sequence lengths. Similarly, Attention with Linear Biases (ALiBi) \citep{press2021alibi} adds a recency bias in the attention mechanism to help the model extrapolate to longer sequence lengths. However, recent work by \cite{dey-btlm} shows ALiBi position embeddings only extrapolate well to $\sim$12\% beyond the trained sequence length for an over-trained model \citep{hoffmann2022chinchilla}. Thus, we propose to extend the sequence extrapolation capabilities of ALiBi position embeddings using position interpolation. Position interpolation with ALiBi extends sequence lengths of models up to 2x the maximum training sequence length while maintaining its original language modelling performance.

\section{Position Interpolation}
Linear position interpolation (PI) was proposed concurrently by \cite{chen-pi} and \cite{kaiokendev-pi} to extend the effective context length of models using rotary position embeddings (RoPE) \citep{su2022roformer}. In this work, we extend position interpolation to also improve the extrapolation capability of models with ALiBi position embeddings, without performing additional pre-training or fine-tuning.

\subsection{ALiBi with Linear Position Interpolation}
In its original implementation, ALiBi adds a bias vector to the query-key dot product. The bias vector is formulated based on predefined slopes for each head, and the positional differences between the queries and the keys. Thus, we can formulate the attention scores for $\text{head}_j$ and $\text{query}_i$ as:
$$\text{softmax}\left(q_iK^T+\underbrace{m_j\cdot[-(i-1),\ldots,-2,-1,0]}_\text{positional bias}\right)$$
To enable ALiBi extrapolation during inference, we propose scaling the slopes dynamically by a factor of $L/L'$ where $L$ is the maximum sequence length observed during training and $L'$ is the extended input sequence length during inference, $m'_j=m_j\left(\frac{L}{L'}\right)$. Noting, we only scale the slopes when $L'> L$ to maintain the previous performance of the model for samples with sequence length smaller than or equal to the training sequence length.

\begin{figure}[h]
    \centering
    \includegraphics[width=\linewidth]{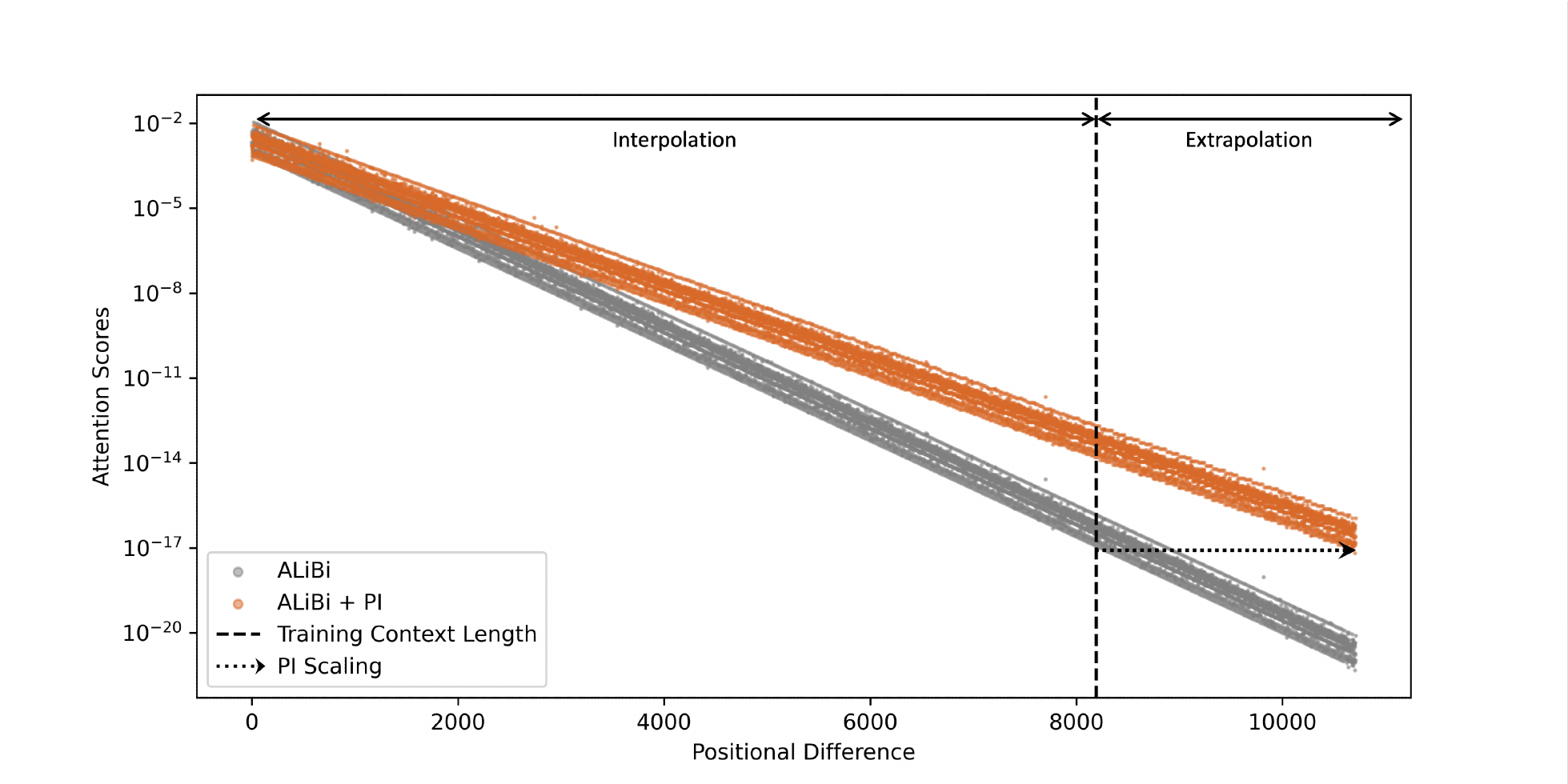}
    \vspace{-6pt}
    \caption{\centering Attention scores in an exemplar attention head (last head of first layer) from BTLM-3B-8K model for a query at position 10.5K.}
    \label{fig-atten}
\end{figure}

In models using RoPE, attention score magnitudes tend to blow up during extrapolation \citep{chen-pi}. Hence, the motivation for using position interpolation with RoPE was to scale down positional distances to the stable and bounded interpolation range seen during training. Contrary to RoPE, we observe ALiBi introduces lower magnitude attention scores for tokens in the extrapolation regime than are seen in the interpolation regime (Figure \ref{fig-atten}). By applying position interpolation, we adjust the ALiBi slope to scale up attention scores and prevent the introduction of lower magnitudes for positional differences beyond the training context length.

\section{Evaluation}
We tested ALiBi with position interpolation using BTLM-3B \citep{dey-btlm} and MPT-7B \citep{mosaic2023mpt7b8k,mosaicmlmpt7b} pre-trained models without fine-tuning.

\subsection{Language Modeling}
To quantify the improvement in extrapolation capability from using position interpolation with ALiBi position embeddings, we follow \cite{Peng-yarn} and measure the average perplexity on 10 documents from Proof-pile \citep{Zhangir-proof-pile} truncated to 16K tokens. 

\begin{figure}[h]
    \centering
    \includegraphics[width=0.8\linewidth]{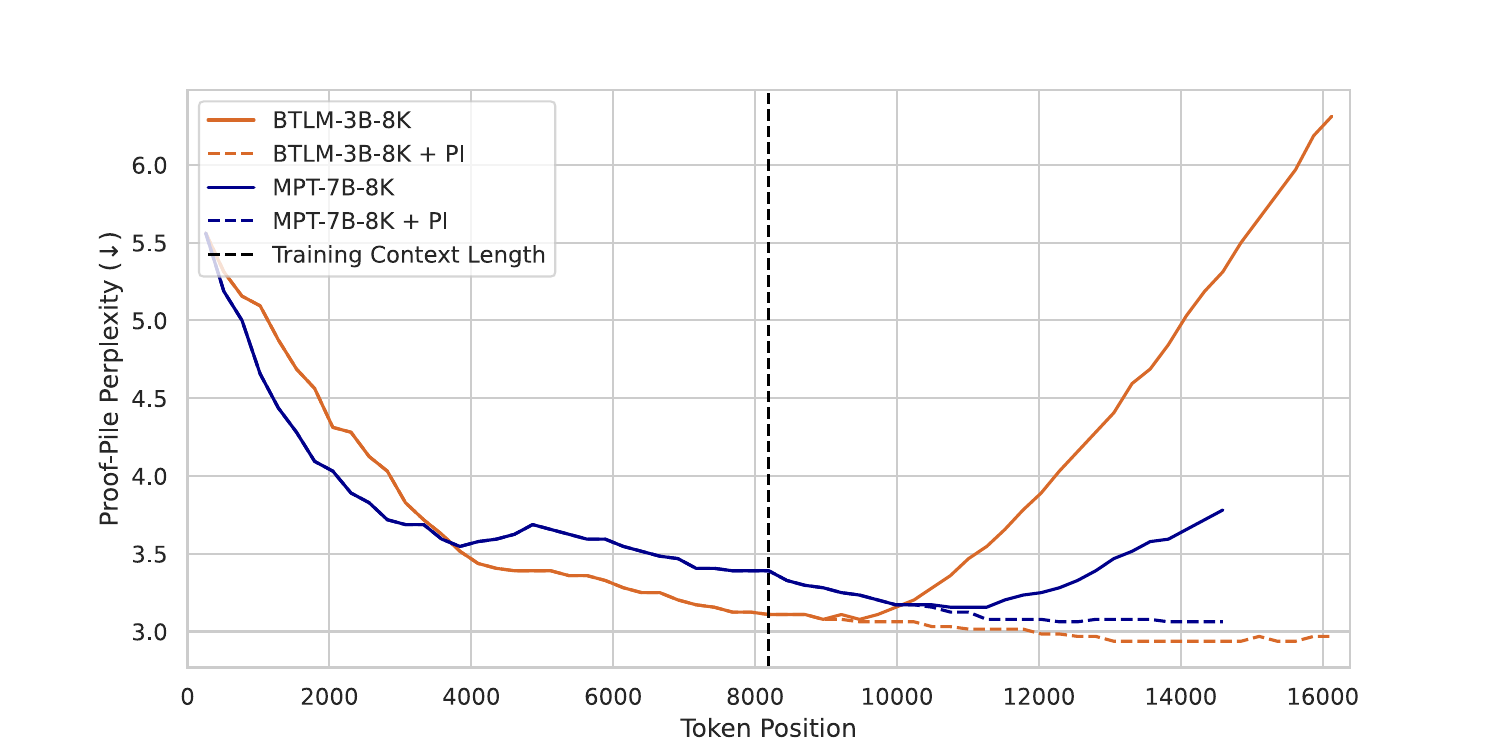}
    \vspace{-6pt}
    \caption{\centering Average perplexity per token position over a set of ten Proof-pile documents for pre-trained models with 8K training context length.}
    \label{fig-lm8k}
\end{figure}

In Figure \ref{fig-lm8k}, we evaluate BTLM-3B-8K and MPT-7B-8K models and observe the baseline models can only extrapolate to 9K-10K context lengths. On the other hand, when using position interpolation these models are able to maintain the same low perplexity for up to at least 16K tokens (2x the training maximum sequence length).

\begin{figure}[h]
    \centering
    \includegraphics[width=0.8\linewidth]{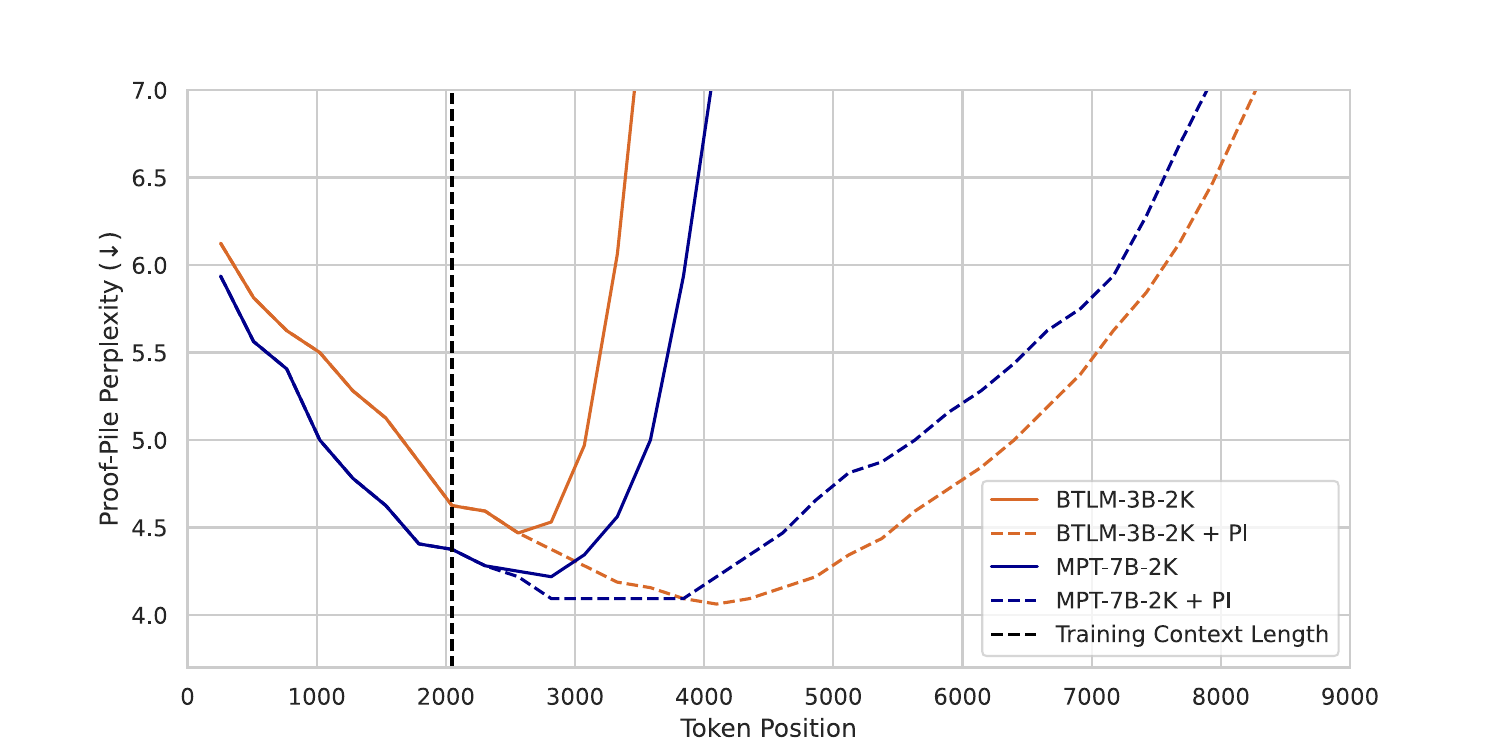}
    \vspace{-6pt}
    \caption{\centering Average perplexity per token position over a set of ten Proof-pile documents for pre-trained models with 2K training context length.}
    \label{fig-lm2k}
\end{figure}

Similarly, in Figure \ref{fig-lm2k} we evaluate BTLM-3B-2K\footnote{The final checkpoint from the 2048 context length training phase.} and MPT-7B-2K and find the baseline models extrapolate to $\sim$2.5K-3K tokens. Position interpolation extends the extrapolation to $\sim$4K tokens before perplexity degradation. It should also be noted that perplexity increases more slowly as token position is increased beyond the training context length.

\subsection{Document Summarization Tasks}
Document summarization is an important application for LLMs that requires inference at long context lengths. We evaluated BTLM-3B-8K up to 16K context lengths on GovReports \citep{huang2021govreport} and QMSum \citep{zhong2021qmsum} summarization tasks. The two tasks require the model to understand long context documents based on samples of government reports and meeting transcripts, respectively. Table \ref{table:scrolls} shows that position interpolation significantly improves BTLM-3B-8K's ability to summarize 16K context lengths, roughly doubling ROUGE scores across tasks.

\begin{table}[h]
\centering
\begin{tabular}{l|lll|lll}
\thickhline
\multirow{2}{*}{Model} & \multicolumn{3}{c|}{QMSum ($\uparrow$)} &  \multicolumn{3}{c}{GovReports ($\uparrow$)}       \\
                 & R-1     & R-2    & R-L     & R-1        & R-2    & R-L     \\
\hline
BTLM-3B-8K & 7.3 & 2.1 & 5.8 & 8.1 & 3.5 & 5.7  \\
BTLM-3B-8K + PI  & \textbf{16.6}  & \textbf{4.7} & \textbf{12.8} & \textbf{14.7} & \textbf{7.0} & \textbf{9.9} \\
\thickhline
\end{tabular}
\caption{ROUGE scores on the QMSum and GovReports long text summarization tasks. We only evaluate samples less than 16,384 tokens in length.}
\label{table:scrolls}
\end{table}

\subsection{Long Range Retrieval Tasks}
\begin{figure}[h]
    \centering
    \includegraphics[width=\linewidth]{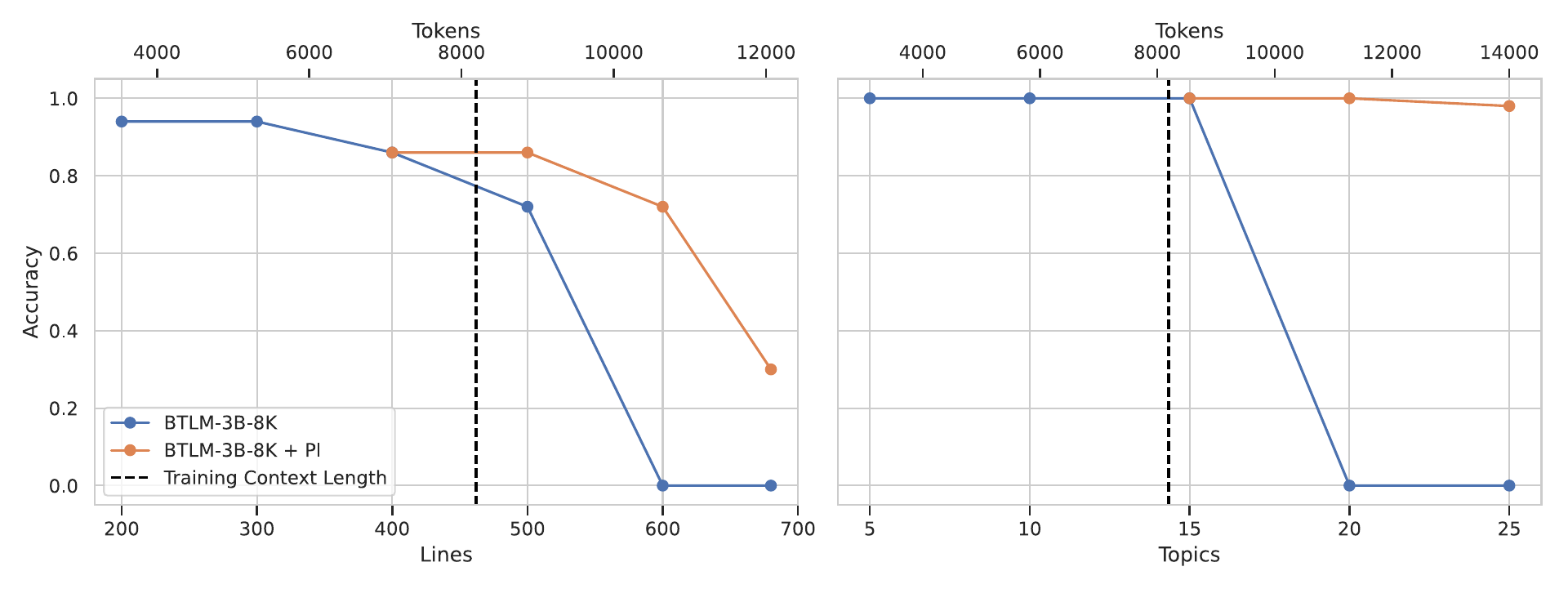}
    \vspace{-6pt}
    \caption{\centering Accuracy on the LongEval line retrieval and topic retrieval tasks.}
    \label{fig-longeval}
\end{figure}

To evaluate long range retrieval capabilities, we use both LongEval tasks \citep{dacheng2023longchat}. The “Coarse-grained Topic Retrieval” task requires models to retrieve the first discussed topic from a long conversation that spans multiple topics, and the “Fine-grained Line Retrieval” task which requires models to precisely retrieve a number from a long document. Figure \ref{fig-longeval} shows that position interpolation greatly improves BTLM-3B-8K's ability to retrieve information from documents longer than seen during training, in both tasks. The line retrieval task being more challenging still suffered from a drop in performance when extending the document length.

\section{Conclusion}
In this work, we extend the position interpolation method to significantly improve the extrapolation capability of models with ALiBi position embeddings. With no additional training, we demonstrate this improvement on long context language modeling, document summarization, and retrieval tasks. As future work, this combination of ALiBi position embeddings and position interpolation could be used to achieve further improvements by fine-tuning models with longer context lengths \cite{chen-pi} than seen during training.

\newpage
\bibliography{main}

\begin{thebibliography}{14}
\providecommand{\natexlab}[1]{#1}
\providecommand{\url}[1]{\texttt{#1}}
\expandafter\ifx\csname urlstyle\endcsname\relax
  \providecommand{\doi}[1]{doi: #1}\else
  \providecommand{\doi}{doi: \begingroup \urlstyle{rm}\Url}\fi

\bibitem[Azerbayev et~al.(2022)Azerbayev, Ayers, and Piotrowski]{Zhangir-proof-pile}
Zhangir Azerbayev, Edward Ayers, and Bartosz Piotrowski.
\newblock {Proof-pile}, 2022.
\newblock URL \url{https://github.com/zhangir-azerbayev/proof-pile}.

\bibitem[Chen et~al.(2023)Chen, Wong, Chen, and Tian]{chen-pi}
Shouyuan Chen, Sherman Wong, Liangjian Chen, and Yuandong Tian.
\newblock {Extending Context Window of Large Language Models via Positional Interpolation}, 2023.
\newblock URL \url{https://arxiv.org/abs/2306.15595}.

\bibitem[Dey* et~al.(2023)Dey*, Soboleva*, Al-Khateeb, Yang, Pathria, Khachane, Muhammad, Chen, Myers, Steeves, Vassilieva, Tom, and Hestness]{dey-btlm}
Nolan Dey*, Daria Soboleva*, Faisal Al-Khateeb, Bowen Yang, Ribhu Pathria, Hemant Khachane, Shaheer Muhammad, Zhiming~(Charles) Chen, Robert Myers, Jacob~Robert Steeves, et~al.
\newblock {BTLM-3B-8K: 7B Parameter Performance in a 3B Parameter Model}, 2023.
\newblock URL \url{https://arxiv.org/abs/2309.11568}.

\bibitem[Dosovitskiy et~al.(2021)Dosovitskiy, Beyer, Kolesnikov, Weissenborn, Zhai, Unterthiner, Dehghani, Minderer, Heigold, Gelly, Uszkoreit, and Houlsby]{dosovitskiy-imageTransf}
Alexey Dosovitskiy, Lucas Beyer, Alexander Kolesnikov, Dirk Weissenborn, Xiaohua Zhai, Thomas Unterthiner, Mostafa Dehghani, Matthias Minderer, Georg Heigold, Sylvain Gelly, et~al.
\newblock {An Image is Worth 16x16 Words: Transformers for Image Recognition at Scale}, 2021.
\newblock URL \url{https://arxiv.org/abs/2010.11929}.

\bibitem[Hoffmann et~al.(2022)Hoffmann, Borgeaud, Mensch, Buchatskaya, Cai, Rutherford, de~las Casas, Hendricks, Welbl, Clark, Hennigan, Noland, Millican, van~den Driessche, Damoc, Guy, Osindero, Simonyan, Elsen, Vinyals, Rae, and Sifre]{hoffmann2022chinchilla}
Jordan Hoffmann, Sebastian Borgeaud, Arthur Mensch, Elena Buchatskaya, Trevor Cai, Eliza Rutherford, Diego de~las Casas, Lisa~Anne Hendricks, Johannes Welbl, Aidan Clark, et~al.
\newblock {An Empirical Analysis of Compute-optimal Large Language Model Training}.
\newblock In \emph{The Conference on Neural Information Processing Systems (NeurIPS)}, 2022.

\bibitem[Huang et~al.(2021)Huang, Cao, Parulian, Ji, and Wang]{huang2021govreport}
Luyang Huang, Shuyang Cao, Nikolaus Parulian, Heng Ji, and Lu~Wang.
\newblock {Efficient Attentions for Long Document Summarization}.
\newblock In \emph{Proceedings of the 2021 Conference of the North American Chapter of the Association for Computational Linguistics: Human Language Technologies}, 2021.
\newblock URL \url{https://aclanthology.org/2021.naacl-main.112}.

\bibitem[kaiokendev(2023)]{kaiokendev-pi}
kaiokendev.
\newblock {Things I’m learning while training superhot}, 2023.
\newblock URL \url{https://kaiokendev.github.io/til#extending-context-to-8k}.

\bibitem[Li* et~al.(2023)Li*, Shao*, Xie, Sheng, Zheng, Gonzalez, Stoica, Ma, and Zhang]{dacheng2023longchat}
Dacheng Li*, Rulin Shao*, Anze Xie, Ying Sheng, Lianmin Zheng, Joseph~E. Gonzalez, Ion Stoica, Xuezhe Ma, and Hao Zhang.
\newblock {How Long Can Open-Source LLMs Truly Promise on Context Length?}, 2023.
\newblock URL \url{https://lmsys.org/blog/2023-06-29-longchat}.

\bibitem[Peng et~al.(2023)Peng, Quesnelle, Fan, and Shippole]{Peng-yarn}
Bowen Peng, Jeffrey Quesnelle, Honglu Fan, and Enrico Shippole.
\newblock {YaRN: Efficient Context Window Extension of Large Language Models}, 2023.
\newblock URL \url{https://arxiv.org/abs/2309.00071}.

\bibitem[Press et~al.(2021)Press, Smith, and Lewis]{press2021alibi}
Ofir Press, Noah~A. Smith, and Mike Lewis.
\newblock {Train Short, Test Long: Attention with Linear Biases Enables Input Length Extrapolation}, 2021.
\newblock URL \url{https://arxiv.org/abs/2108.12409}.

\bibitem[Su et~al.(2022)Su, Lu, Pan, Murtadha, Wen, and Liu]{su2022roformer}
Jianlin Su, Yu~Lu, Shengfeng Pan, Ahmed Murtadha, Bo~Wen, and Yunfeng Liu.
\newblock {RoFormer: Enhanced Transformer with Rotary Position Embedding}, 2022.
\newblock URL \url{https://arxiv.org/abs/2104.09864}.

\bibitem[Team(2023{\natexlab{a}})]{mosaic2023mpt7b8k}
MosaicML~NLP Team.
\newblock {Announcing MPT-7B-8K: 8K Context Length for Document Understanding}, 2023{\natexlab{a}}.
\newblock URL \url{https://www.mosaicml.com/blog/long-context-mpt-7b-8k}.

\bibitem[Team(2023{\natexlab{b}})]{mosaicmlmpt7b}
MosaicML~NLP Team.
\newblock Introducing mpt-7b: A new standard for open-source, commercially usable llms, 2023{\natexlab{b}}.
\newblock URL \url{www.mosaicml.com/blog/mpt-7b}.

\bibitem[Zhong et~al.(2021)Zhong, Yin, Yu, Zaidi, Mutuma, Jha, Awadallah, Celikyilmaz, Liu, Qiu, and Radev]{zhong2021qmsum}
Ming Zhong, Da~Yin, Tao Yu, Ahmad Zaidi, Mutethia Mutuma, Rahul Jha, Ahmed~Hassan Awadallah, Asli Celikyilmaz, Yang Liu, Xipeng Qiu, and Dragomir Radev.
\newblock {QMSum: A New Benchmark for Query-based Multi-domain Meeting Summarization}.
\newblock In \emph{Proceedings of the 2021 Conference of the North American Chapter of the Association for Computational Linguistics: Human Language Technologies}, 2021.
\newblock URL \url{https://aclanthology.org/2021.naacl-main.472}.

\end{thebibliography}
\bibliographystyle{tmlr}

\end{document}